\documentclass[letterpaper, 10 pt, journal, twoside]{ieeetran}

\pdfoutput=1

\usepackage{amsmath,amssymb,amsfonts,leftidx,setspace}
\usepackage{bm}
\usepackage{url} 
\usepackage{hyperref} 
\usepackage{algpseudocode}
\usepackage[linesnumbered, ruled]{algorithm2e}
\usepackage{graphicx}
\usepackage{textcomp}
\usepackage{xcolor}
\usepackage[most]{tcolorbox}
\usepackage{multirow}
\usepackage[normalem]{ulem}
\usepackage{cancel}
\usepackage{tikz}
\usepackage{tablefootnote}
\usepackage{threeparttable}
\usepackage{cite}
\usepackage[section]{placeins} 

\usepackage{dblfloatfix}   
\usepackage{geometry}
\newcommand{\SE}{\text{SE}}
\newcommand{\SO}{\text{SO}}
\newcommand{\II}{\mathbb{I}}
\newcommand{\IR}{\mathbb{R}}

\newcommand{\xx}{{\bf x}}
\newcommand{\pp}{{\bf p}}

\newcommand{\nn}{{\bf n}}
\newcommand{\xt}{{\bf t}}

\newcommand{\ie}{{\it i.e.,}}

\newcommand{\PP}{\text{P}}

\author{Xiaoli Wang$^{\dagger1}$, Sipu Ruan$^{\dagger1}$, Xin Meng$^{1}$, and Gregory S. Chirikjian$^{*1,2}$

\thanks{Manuscript received: February, 18, 2025; Revised July, 20, 2025; Accepted August, 10, 2025. 
This paper was recommended for publication by Editor O. Stasse upon evaluation of the Associate Editor and Reviewers' comments.
This work was supported by NUS Startup grants A-0009059-02-00, A-0009059-03-00, CDE Board account E-465-00-0009-01, and National Research Foundation, Singapore, under its Medium Sized Centre Programme - Centre for Advanced Robotics Technology Innovation (CARTIN), subaward A-0009428-08-00. }

\thanks{$^1$Xiaoli Wang, Sipu Ruan, Xin Meng, and Gregory S. Chirikjian are with Department of Mechanical Engineering, National University of Singapore, Singapore. (\texttt{email: wangxiaoli@u.nus.edu, rsp01@163.com, mengxin@u.nus.edu, mpegre@nus.edu.sg})}

\thanks{$^2$Gregory S. Chirikjian is also with Department of Mechanical
Engineering, University of Delaware, USA. (\texttt{email: gchirik@udel.edu})}

\thanks{$^\dagger$The authors contribute equally. $^*$Corresponding author. }

\thanks{Digital Object Identifier (DOI): see top of this page.}}

\title{Enhanced Probabilistic Collision Detection for Motion Planning Under Sensing Uncertainty }

\begin{document}

\maketitle

\begin{abstract}
Probabilistic collision detection (PCD) is essential in motion planning for robots operating in unstructured environments, where considering sensing uncertainty helps prevent damage. Existing PCD methods mainly use simplified geometric models and address only position estimation errors. This paper presents an enhanced PCD method with two key advancements: (a) using superquadrics for more accurate shape approximation and (b) accounting for both position and orientation estimation errors to improve robustness under sensing uncertainty. Our method first computes an enlarged surface for each object that encapsulates its observed rotated copies, thereby addressing the orientation estimation errors. Then, the collision probability is formulated as a chance-constraint problem that is solved with a tight upper bound. Both steps leverage the recently developed closed-form normal parameterized surface expression of superquadrics. Results show that our PCD method is twice as close to the Monte-Carlo sampled baseline as the best existing PCD method and reduces path length by $30\%$ and planning time by $37\%$, respectively. A Real2Sim2Real pipeline further validates the importance of considering orientation estimation errors, showing that the collision probability of executing the planned path is only $2\%$, compared to $9\%$ and $29\%$ when considering only position estimation errors or no errors at all.

\end{abstract}


\section{Introduction}

\begin{figure}[tb]
\centering
\includegraphics[width=0.9\linewidth]{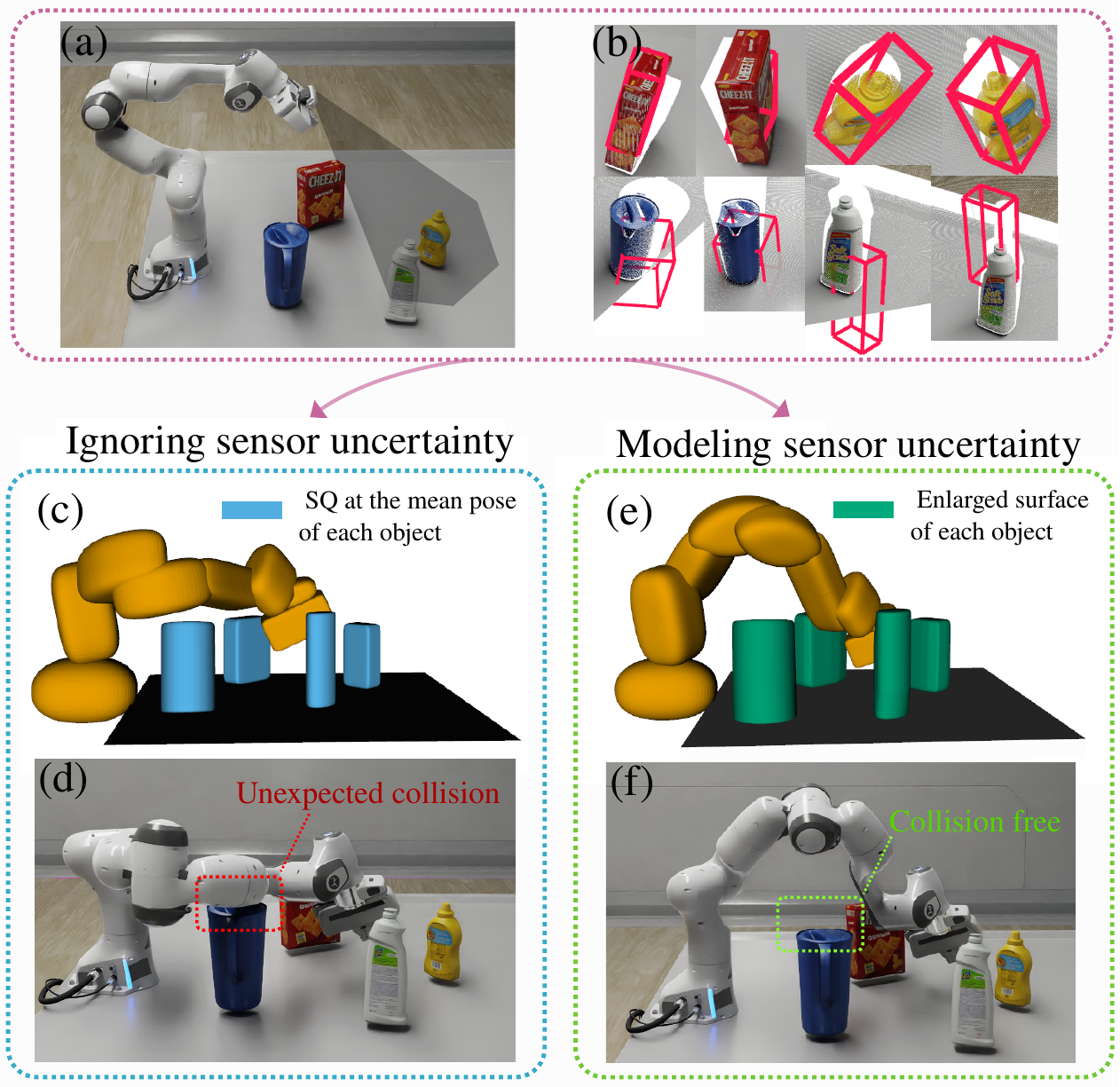}
\caption{Comparison of motion planning results with and without taking into account sensing uncertainty. (a-b) Pose estimates of each object using an existing method in Isaac Sim \cite{wen2024foundationpose}. (c) Superquadric representation of each robot link at its ground truth pose (yellow) and of each object at its estimated mean pose (blue), where the robot configuration is justified as valid by a deterministic collision checker \cite{ruan2022collision}. (d) Snapshot of a collision between the robot and an object due to unmodeled pose uncertainty. (e) The enlarged surface that encapsulates many rotated copies of the objects due to orientation errors in pose estimates (green), and the robot configuration that is justified as valid by our probabilistic collision checker. (f) Corresponding snapshot of the robot showing no collision with objects. }
\label{fig:with/without-pcd}
\end{figure}

\IEEEPARstart{C}{ollision} detection is essential to motion planning, which helps to prevent robots from colliding with their surroundings. Although traditional collision detection methods have been developed for decades, they usually assume perfect knowledge of the states of robots and environments \cite{ruan2022collision}. This assumption does not apply in most real-world applications, especially for service robots with a high degree of freedom (DOF) manipulating objects in domestic settings. In such cases, robots will need to interact with objects closely, but the pose estimation of objects is often affected by occlusions and sensor inaccuracies. The sensing uncertainty may cause unexpected collisions that lead to a failure of manipulation and damage the robot and the surroundings. An example is shown in Fig. \ref{fig:with/without-pcd} (c-d).

Compared with deterministic collision detection, probabilistic collision detection (PCD) takes into account the sensing uncertainty and calculates the collision probability between two bodies in a single query. When a PCD is used as the collision checker in a motion planner for a robot, it gives the collision probability between each link of the robot and each environmental object. The PCD outcome enables the motion planner to quantify the level of safety, ensuring that the overall collision probability between the robot and objects along the planned path is lower than a given threshold. 

However, existing PCD methods usually use a simplified geometric model (e.g., points or ellipsoids) for the robot and environmental objects and only account for position estimation errors \cite{zhu2019chance, liu2023tight, thomas2022safe, park2020efficient}. These assumptions may lead to less robustness when complex environmental objects cannot be accurately represented by simple geometries and when their orientation estimation errors cannot be ignored. 

To address these limitations, this paper presents an enhanced PCD method with two key advancements. First, it supports using a broad class of geometric models, specifically superquadrics, to approximate the true shape of environmental objects accurately. Superquadrics are a family of geometric shapes defined by five parameters resembling ellipsoids and other quadrics, as introduced in Section \ref{subsec:pre:normal-surface}. This flexibility allows them to accurately approximate the standard collision geometric models (e.g., bounding box or cylinder) or represent details of complex objects. An example of the superquadrics representation for each robot link and environmental object is shown in Fig. \ref{fig:with/without-pcd} (c). We applied an existing method for the superquadric-based shape fitting \cite{vaskevicius2017revisiting}, and compared the relative goodness of model fit using superquadrics and ellipsoids in the Section \ref{subsec:benchmark:model-fit}. 
The second key advancement is that the proposed method accounts for object position and orientation estimation errors, increasing robustness against sensing uncertainty. More specifically, we proposed to use an enlarged surface that encapsulates many rotated copies of the objects due to orientation errors in pose estimates, as shown in Fig. \ref{fig:with/without-pcd} (e). Overall, these improvements enhance the reliability of PCD in unstructured environments.

We conducted a benchmark on PCD methods and found that our method considerably refines the accuracy of collision probability estimation. In addition, we examined how the accuracy and computation time of PCD methods influence the performance of motion planners. Results show that with our PCD method, planners find more efficient paths within less planning time, especially in cluttered scenes. Furthermore, we designed a Real2Sim2Real pipeline to assess the necessity of considering the orientation estimation errors of objects. We compared three different kinds of planners: 1) deterministic; 2) PCD including position errors; 3) PCD with both position and orientation errors. Results show that when both the position and orientation uncertainties are considered in motion planning, the collision probability during execution consistently exhibits the lowest risk.  

The major contributions of this work are:
\begin{itemize}
\item Whereas previous methods only handle positional uncertainty, we present a novel method to handle orientation uncertainty using an enlarged parametric surface that encapsulates rotated copies of the object.  
\item Tight approximation for collision probability: An efficient linear chance constraint is proposed that gives a tight upper bound for collision probability under position errors.
\item Real2Sim2Real evaluation pipeline: A new pipeline that quantifies the reliability of planned paths by simulating their execution under real-world sensing uncertainties. 
\end{itemize}

\section{Related Work}
Recently, PCD has been developed to account for uncertainties in robot controllers and environment sensing to ensure the safety of robots operating in the real world, such as drones and autonomous cars \cite{zhu2019chance}\cite{dawson2020provably}. However, PCD for high-DOF robots is particularly challenging as they often need to interact closely with objects in cluttered scenes, such as during pick-and-place tasks, so even small errors in depth or pose estimation can add up and raise the chance of undetected collisions.


Early PCD methods, such as Monte Carlo-based approaches, compute the collision probability accurately but are computationally demanding, limiting their practical use in real-time applications \cite{lambert2008fast}. Some PCD methods that have closed-form solutions are fast to compute and are suitable for high-speed robots \cite{du2011probabilistic, park2018fast, zhu2019chance}. Nevertheless, they only support using simple geometric primitives (e.g., points, spheres, or ellipsoids) to represent robots and environmental objects, which can give a conservative result if the objects cannot be accurately represented. Moreover, they only consider the position estimation errors of objects. Although some PCD methods support convex complex geometric models (e.g., mesh), their performance either depends on the surface complexity of the model \cite{park2020efficient} or needs to be iteratively improved \cite{dawson2020provably}. Nevertheless, they only support the position estimation errors of objects. Learning-based methods use the point cloud of the environment and do not assume a probabilistic model per object, but the computation time is too expensive, and retraining is required for new robots. \cite{quintero2024stochastic}. 

Despite these advancements, current PCD methods still face limitations in accurately modeling complex shapes and handling combined position and orientation uncertainties. This work addresses these gaps by introducing a robust PCD method that leverages superquadrics for improved shape approximation and integrates position and orientation uncertainties for enhanced robustness in real-world applications.

\section{Preliminary}

\subsection{Probabilistic collision detection}

The collision condition for two convex bodies $S_1$, $S_2$ with exact poses can be written as 
\begin{equation*}
\text{if}\,  S_1 \cap S_2 \neq \varnothing \Leftrightarrow \pp_2 \in S_1 \oplus (-S_2) \, .
\end{equation*}
where\hspace{0.2em}$S_1\hspace{-0.2em}\oplus\hspace{-0.2em}(-S_2)$\hspace{0.2em}is the Minkowski sum and $-S_2$ is the reflection of $S_2$ about its center $\pp_2$. 

If $S_i$ is moved by $g_i=(R_i, \xt_i) \in \SE(3)$, we denote the resulting body as $S_i^{g_i}=R_iS_i+\xt_i$. If the poses $g_1$ and $g_2$ are inaccurate, the collision status is probabilistic and can be written as $\PP(S_1^{g_1}, S_2^{g_2}) = P(S_1^{g_1} \cap S_2^{g_2} \neq \varnothing)$.

\subsection{Collision probability under position sensing uncertainty}
\label{subsec:preliminary:probability under position estimation uncertainty}
Previous PCD methods only consider inaccurate translation estimates $\xt_i$ of objects. The inaccurate point set is $S_i^{\xt_i} = S_i + \xt_i$, and the collision condition becomes:
\begin{equation}
    \begin{aligned}
  &(S_1 + \xt_1)\,\cap\,(S_2 + \xt_2)\neq\emptyset \\
  &\Rightarrow 
   \exists\,\mathbf{s}_1\in S_1,\,\mathbf{s}_2\in S_2:\,(\mathbf{s}_2 + \xt_2) - (\mathbf{s}_1 + \xt_1) = \mathbf{0} \, ,\\
  &\Leftrightarrow \,
   \xt_2 - \xt_1 = \mathbf{s}_1 - \mathbf{s}_2, \, \Leftrightarrow \,
   \xt_2 - \xt_1 \in S_1 \oplus (-S_2).
\end{aligned}
\end{equation}
The collision probability $\PP(S_1^{\xt_1}, S_2^{\xt_2})$ can be written as the integral:
\begin{equation}
\label{eq:pre:pcd-position}
\int_{\IR^3} \hskip -0.05in \int_{\IR^3} \hskip -0.08in \iota(\xt_2-\xt_1 \in S_1\oplus(-S_2)) \rho_1(\xt_1)\rho_2(\xt_2)d\xt_1d\xt_2 \, ,
\end{equation}
where $\iota(\cdot)$ is an indicator function that returns $1$ when the input condition is true and $0$ otherwise. $\rho(\cdot)$ is the shorthand for the probability density function (pdf) defined by the argument. 

The position error is usually modeled as zero-mean Gaussian distributed, i.e.,\hspace{0.1em}$\xt_i\hspace{-0.2em}\sim\hspace{-0.2em}\mathcal{N}(\mathbf{0}, \Sigma_i)$, where $\Sigma$ is the covariance.
To make the formulas in this work concise and tight, we move the objects $S_1$ and $S_2$ to the global origin, and then $S_1\oplus(-S_2)$ is centered at the global origin. Thus, the mean of the zero-mean additive position error is $\xt_i\hspace{-0.2em}\sim\hspace{-0.2em}\mathcal{N}(\pp_i, \Sigma_i)$, where $\pp_i$ is the center position of object $S_i$.

\subsection{Linear chance constraint approximation}
The collision probability in (\ref{eq:pre:pcd-position}) has no closed-form solution, and its numerical solution is computationally expensive. Instead, (\ref{eq:pre:pcd-position}) is usually written as a chance-constraint $\PP(S_1^{\xt_1}, S_2^{\xt_2}) \leq \delta$, where $\delta$ is an upper bound for the collision probability.

When $\xt$ is Gaussian distributed, $\delta$ has a closed-form solution \cite{blackmore2011chance}. The idea is that the integration of a Gaussian distributed variable $\xt$ inside a linear half-space can be transformed as a cumulative distribution function (cdf) of a new 1D Gaussian variable \cite{blackmore2011chance}:
\begin{equation*}
\int_{\mathbf{a}^T\xt-b<0} \rho(\xt_{21}; \pp_{21}, \Sigma_{21}) d\xt_{21} = \int_{y<0} \rho(y; p_y, \sigma_y) dy \, ,
\end{equation*}
where $\xt_{21}$ is the relative position error $\xt_{21}=\xt_2-\xt_1$, $\mathbf{a}^T\xt-b<0$ defines a linear half-space with the normal $\mathbf{a}$ and the constant $b$, and $y=\mathbf{a}^T\xt-b$ is a 1D variable that follows Gaussian distribution $\mathcal{N}(\mathbf{a}^T\pp_{21}-b, \mathbf{a}^T\Sigma_{21}\mathbf{a})$. The cdf result $F_y(0)$ can be easily found by the look-up table. 

This means that if the integration region $S_1\oplus(-S_2)$ in (\ref{eq:pre:pcd-position}) is fully contained by a linear half-space, the collision probability in (\ref{eq:pre:pcd-position}) can be bounded by
\begin{equation}
\label{eq:pre:linear-chance-constraint-approximation}
\PP(S_1^{\xt_1}, S_2^{\xt_2}) = \int_{S_1\oplus(-S_2)} \rho(\xt; \pp_{21}, \Sigma_{21}) d\xt < F_y(0) \, .
\end{equation}
The accuracy of the approximation depends on the choice of the linear half-space. Previous methods choose a naive plane with a closed-form solution by using the normalized $\pp_{21}$ of the relative position error as the normal $\mathbf{a}$ when objects are spheres or ellipsoids \cite{zhu2019chance} \cite{liu2023tight}. It is named as \textit{lcc-center} for reference. 

Although \textit{lcc-center} is computationally efficient, the geometric models are limited, and the result using such a plane can be conservative. In this work, we propose a linear half-space that gives a more accurate approximation for the collision probability than \textit{lcc-center}. 

\subsection{Normal-parameterized superquadric surface}
\label{subsec:pre:normal-surface}
The implicit function of a superquadric is given as: $$
\Psi(\xx) = \left( \left(\frac{x_1}{a_1}\right)^{\frac{2}{\epsilon_2}} + \left(\frac{x_2}{a_2}\right)^{\frac{2}{\epsilon_2}} \right)^{\frac{\epsilon_2}{\epsilon_1}} + \left(\frac{x_3}{a_3}\right)^{\frac{2}{\epsilon_1}} = 1 \, 
$$
A superquadric is a flexible model defined by only five parameters, where the semi-axes $\mathbf{a}=[a_1, a_2, a_3]$ define the size of the superquadric along the principal axes and the epsilons $\epsilon_1, \epsilon_2$ control the shape, where $\epsilon_1 , \epsilon_2 \in (0,2)$ ensure the convexity. 
When $\epsilon_1=\epsilon_2=1$, superquadrics degenerate to be ellipsoids. The closed-form normal-parameterized surface expression $\xx=\xx(\nn)\in\partial S$ can be found in \cite{ruan2022collision} and references therein.

For two superquadrics $S_1$ and $S_2$, given the normal parameterization of $\xx_i(\nn_i)$, the point on the Minkowski sum boundary $\xx_\Sigma\hspace{-0.2em}\in\hspace{-0.2em}\partial[S_1\hspace{-0.2em}\oplus\hspace{-0.2em}(-S_2)]$ can be easily found as \cite{ruan2022collision}
$$\xx_\Sigma(\nn_1)=\xx_1(\nn_1)-\xx_2(-\nn_1)$$ because the normals at the contact points of two bodies are anti-parallel, i.e., $\nn_1\hspace{-0.2em}=\hspace{-0.2em}-\nn_2$, where the normal at $\xx_\Sigma$ is the same as the normal at $\xx_1$ of $S_1$.

If applying the rotations $R_1, R_2 \in SO(3)$ to objects, $\xx_\Sigma \in \partial[R_1S_1\oplus(-R_2S_2)]$ can be written as $\xx_\Sigma(\nn) = R_1\xx_1(R_1^T\nn)-R_2\xx_2(-R_2^T\nn)$, where $\nn$ is the normal at $\xx_\Sigma$, where $R_1^T\nn$ and $R_2^T\nn$ are normals of points at the original body boundary $\partial S_1$ and $\partial S_2$, respectively. 

\begin{figure}[t]
\centering
\includegraphics[width=0.85\linewidth]{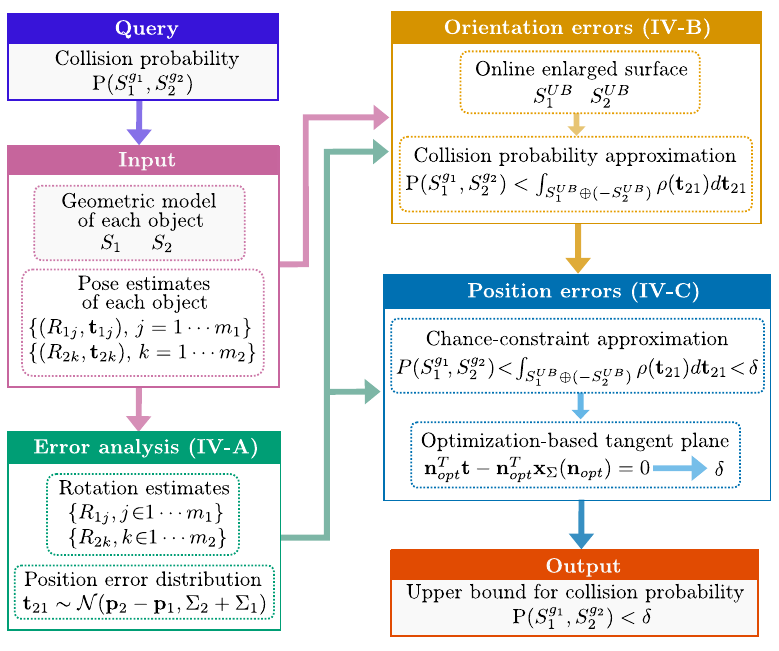}
\caption{Proposed probabilistic collision detection method for approximating collision probability between two objects under pose sensing errors. }
\label{fig:method:pose_estimate_errors}
\end{figure}

\section{Probabilistic Collision Detection}

Let $S_1$ and $S_2$ be two rigid bodies. For each body $S_i$, it has a set of pose estimates $\{g_{ij}=(R_{ij},\xt_{ij}),\, j=1\cdots m_i\}$, with \(R_{ij}\in\mathrm{SO}(3)\) and \(\xt_{ij}\in\mathbb R^3\). Due to the imperfect perception module, the pose estimation is inaccurate. The collision probability $\PP(S_1^{g_1}, S_2^{g_2})$ under pose estimation errors can be written as:
\begin{equation}
\label{eq:method:pose-collision-probability}
 \int_{SE(3)} \int_{SE(3)}
  \iota(S_1^{g_1}\cap S_2^{g_2} \neq \varnothing) \rho_1(g_1)\rho_2(g_2) dg_1 dg_2 \, ,
\end{equation}
where $g_i$ represents the randomness of the pose of object $S_i$, distributed according to the density $\rho_i(g_i)$. However, (\ref{eq:method:pose-collision-probability}) has no closed-form solution. The proposed method for approximating the collision probability is shown in Fig. \ref{fig:method:pose_estimate_errors}.

\subsection{Error analysis}

This work assumes that the position and orientation estimates are independent, i.e., $\rho_i(g_i)=\rho_i(\xt_i)\rho_i(R_i)$. 

The position error distribution $\rho_i(\xt_i)$, i.e., $\xt_i\sim \mathcal{N}(\pp_i, \Sigma_i)$, can be computed from $\xt_{ij}$ of the pose estimates. Because the position estimates between $S_1$ and $S_2$ are independent, the distribution of the relative position error $\xt_{21}=\xt_2-\xt_1$ is $\xt_{21} \sim \mathcal{N}(\pp_{21}, \Sigma_{21})$, where $\pp_{21} = \pp_2 - \pp_1$ and $\Sigma_{21} = \Sigma_2 + \Sigma_1$. 

Unlike the position error, the probabilistic model $\rho_i(R_i)$ is not explicitly computed. Given the definition of the mean rotation $R_i$ of the set of rotations $\{R_{ij},j\hspace{-0.2em}\in\hspace{-0.3em}1\cdots m\}$ as \cite{ackerman2013probabilistic}, 
$$ \sum_{j=1}^m \log(R_i^TR_{ij})=\mathbb{O} \, ,$$ 
where $\log(\cdot)\colon\hspace{-0.2em}\mathrm{SO}(3)\hspace{-0.2em}\to\hspace{-0.2em}\mathfrak{so}(3)$ is the matrix logarithm. The mean rotation $R_i$ can be solved iteratively.

\subsection{Enlarged surface to handle orientation uncertainties}
\label{subsec:method:enlarged-surface}

To handle the orientation uncertainty, we choose to expand the surface of $S_i$ so that the enlarged surface encapsulates its rotated copies ${R_{ij}}$. This is inspired by the geometric-based PCD method for position errors \cite{dawson2020provably}. Denoted the enlarged surface of $S_i$ as $S_i^{UB}$. 

\begin{figure}[tb]
\centering
\includegraphics[width=0.9\linewidth]{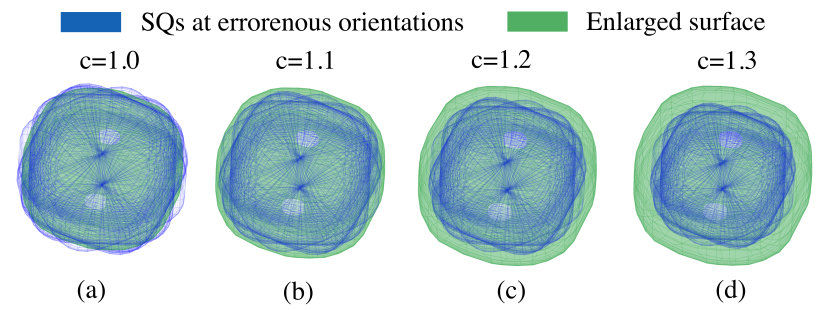}
\caption{The enlarged surfaces of $S_i^{UB}$\hspace{-0.2em} with different scaling constants $c$. The boundary lines of the rotated copies are in blue, and the surface and boundary lines of the enlarged surface of $S_i^{UB}$\hspace{-0.2em} are in green. This work chooses an empirical value $c=1.2$ for the real-world experiment. }
\label{fig:method:SO3}
\end{figure}

To find the enlarged surface of $S_i^{UB}$, we propose a closed-form normal-parameterized expression of the form:
\begin{equation}
\label{eq:method:SO3}
\xx_i^{UB}(\nn_i) = \frac{c}{m} \left( \sum_{j=1}^{m} R_{ij} \xx_i(R_{ij}^T\nn_i) \right) \, ,
\end{equation}
where $\xx_i^{UB}(\nn_i) \in \partial S_i^{UB}$, and $\nn_i$ is the normal of $\partial S_i^{UB}$ at $\xx_i^{UB}$. $m$ is the number of rotation estimates. $c$ is a scaling constant that can be tuned based on the level of sensing uncertainty in the perception module.
(\ref{eq:method:SO3}) can be viewed as the Minkowski sum of $m$ rotated copies of $S_i$ scaled down by $m$ to result in the ‘average body’ that reflects the contribution of the rotated copies, and scaled up by a constant $c$ to ensure encapsulation. 
If the object has an exact orientation, the enlarged surface will be the linearly scaled version of itself. 
In this work, we choose an empirical value of $c=1.2$. An example of scaling the enlarged surface with different scaling constants is shown in Fig. \ref{fig:method:SO3}. The enlarged surface is built online for collision checking. 

A key property of the encapsulating surface constructed from the rotated copies is that it retains the dominant shape characteristics when the camera is accurate. However, if the perception results are too noisy and the rotation distribution is highly diffuse, the enlarged surface would degenerate to a sphere. But in such a case, no method would perform well.

The Minkowski sum boundary $\xx_\Sigma^{UB}(\nn) \in \partial[S_1^{UB}\oplus(-S_2^{UB})]$ can be written as:
\begin{equation}
\label{eq:method:enlarged-minksum}
\begin{aligned}
\xx_\Sigma^{UB}(\nn) =\xx_1^{UB}(\nn) - \xx_2^{UB}(-\nn) \, .
\end{aligned}
\end{equation}
If one body $S_i$ has exact orientation, such as the link of the table-fixed robotic arm, $R_{ij}$ is set to the exact orientation. 

With $S_1^{UB}$ and $S_2^{UB}$, we get an inequality of collision probability (\ref{eq:method:pose-collision-probability}) as:
\begin{equation}
 \label{eq:method:pcd-inequality-orientation}
\PP(S_1^{g_1}, S_2^{g_2}) < \int_{\IR^3} \iota(\xt_{21} \in S_1^{UB} \oplus (-S_2^{UB})) \rho(\xt_{21})d\xt_{21} \, .
\end{equation}
Now, the collision probability under pose sensing uncertainty can be approximated by linearizing the chance constraint problem as shown in (\ref{eq:pre:linear-chance-constraint-approximation}).

\subsection{Position estimation errors}

\begin{figure*}[tb]
\centering 
\includegraphics[scale=0.58]{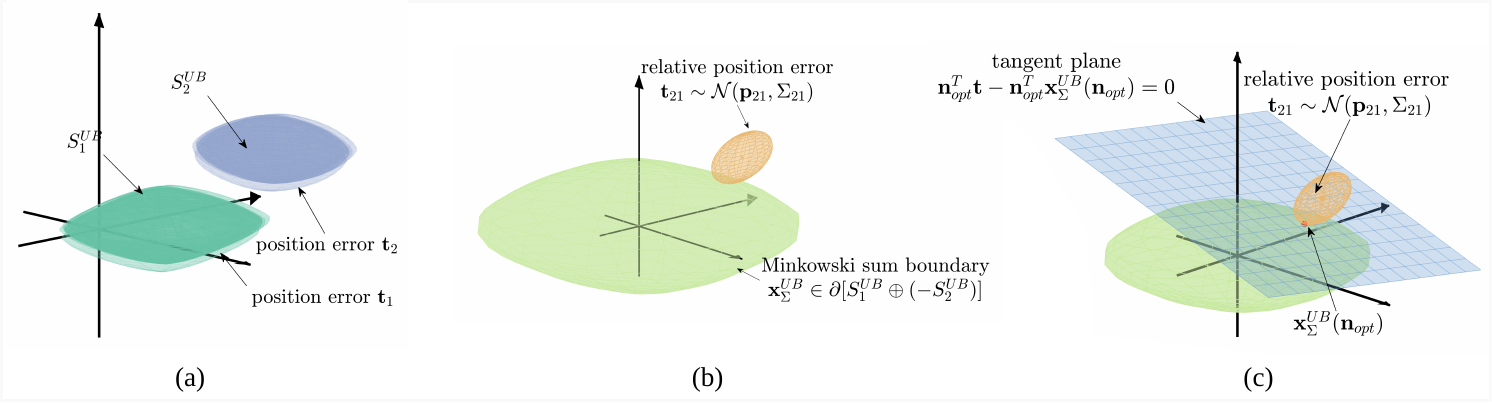}
\caption{Illustration of the linear chance constraint method \textit{\textit{lcc-tangent}}. (a) The two enlarged surfaces $S_1^{UB}$ and $S_2^{UB}$ of two superquadrics $S_1$ and $S_2$, and the translated copies of $S_1^{UB}$ and $S_2^{UB}$ with positions sampled based on the distributions of position errors $\xt_1$ and $\xt_2$; (b) the Minkowski sum boundary $\xx_\Sigma^{UB} \in \partial[S_1^{UB} \oplus (-S_2^{UB})]$, and the ellipsoidal level surface of the relative position error $\xt_{21}=\xt_2-\xt_1$, where $\xt_{21}\sim \mathcal{N}(\pp_{21}, \Sigma_{21})$; (c) the optimized tangent half-space in the untransformed space based on (\ref{eq:method:least-squares}).}
\label{fig:method:half-plane}
\end{figure*}

The quantity in (\ref{eq:method:pcd-inequality-orientation}) can be further bounded as a chance-constraint problem:
\begin{equation}
 \label{eq:method:pcd-lcc}
\int_{\IR^3} \iota(\xt_{21} \in S_1^{UB} \oplus (-S_2^{UB})) \rho(\xt_{21})d\xt_{21} < \delta \, ,
\end{equation}
where $\delta$ is an upper bound and can be solved in closed form. As shown in (\ref{eq:pre:linear-chance-constraint-approximation}), the accuracy of the upper bound $\delta$ for $\PP(S_1^{g_1}, S_2^{g_2})$ depends on the choice of the linear half-space. 

For a Gaussian-distributed position error $\xt_{21}$, the confidence level surface is an ellipsoid, with higher pdf values closer to the center. If the chosen plane is tangent to both the Minkowski sum boundary and the confidence level surface, the resulting half-space contains less of the high pdf region. This provides a better linearization for the chance-constrained PCD problem than the previous \textit{lcc-center}.

To find the tangent half-space, it is important to make the plane tangent to the confidence level surfaces. Because a confidence level surface is an ellipsoid (see Fig. \ref{fig:method:half-plane} (b)), an easy way to find its tangent plane is to transform the ellipsoid into a sphere. This is done by applying the linear transformation $\Sigma_{21}^{-1/2}$ to the whole space.  The transformed position error distribution becomes $\xt_{21}' \sim \mathcal{N}(\pp_{21}', \II)$, where $\pp_{21}'=\Sigma_{21}^{-1/2}\pp_{21}$. Given two enlarged superquadrics, their Minkowski sum region $\xx_\Sigma^{UB}$ before the linear transformation is shown in the green region in Fig. \ref{fig:method:half-plane} (b). $\xx_\Sigma^{UB}$ after the transformation can be calculated as $\xx_\Sigma^{UB'} = \Sigma_{21}^{-1/2} \xx_\Sigma^{UB}$, which still has the closed-form solution and remains convex. 

The problem of finding the tangent half-space is formulated to find a tangent plane on $\xx_\Sigma^{UB'}(\nn)$ with the shortest distance to the center $\pp_{21}'$. This can be formulated as a nonlinear least squares optimization:
\begin{equation}
\label{eq:method:least-squares}
\min_\psi \frac{1}{2}||\pp_{21}'-\xx_\Sigma^{UB'}(\nn(\boldsymbol{\psi}))||_2^2 \, ,
\end{equation}
which is solved by the trust region algorithm. 

Because the optimization in (\ref{eq:method:least-squares}) is not convex, it does not guarantee the global minimum and is sensitive to the initial value $\psi_0$. Inspired by \cite{ruan2022collision}, this work uses the angular parameter of the center of the position error $\pp_{21}'$ as viewed in the body frame of $S_1$ with the mean rotation $R_1$, denoted as $ \leftidx{^1}\pp_{21}'=R_1^T\pp_{21}'$. For 3D cases, $\psi_0$ equals to:
$$ [ \text{atan2}\left( \leftidx{^1}\pp_{21,3}', \sqrt{(\leftidx{^1}\pp_{21,1}')^2 + (\leftidx{^1}\pp_{21,2}')^2} \right), \text{atan2}\left(\leftidx{^1}\pp_{21,2}', \leftidx{^1}\pp_{21,1}'\right) ] $$
After finding the optimized value $\psi_{\text{opt}}$, the normal $\mathbf{a}$ and constant $b$ of the tangent half-space in the untransformed space are $\mathbf{a} = \nn_{\text{opt}}(\psi_{opt})$ and $ b = \nn_{\text{opt}}^T \xx_\Sigma^{UB}(\nn_{\text{opt}})$, respectively. Substituting $\mathbf{a}$ and $b$ into (\ref{eq:pre:linear-chance-constraint-approximation}), the accurate approximation can be calculated, named as \textit{\textit{lcc-tangent}} for reference.

\subsection{Hierarchical PCD}
Although the proposed \textit{\textit{lcc-tangent}} gives a tight approximation for the collision probability, the existing \textit{lcc-center} for ellipsoids computes faster and can be helpful to screen out low collision probability regions quickly. Therefore, this work proposes a hierarchical PCD method, named \textit{h-lcc}, to balance between computation time and better estimation. 

Here the idea of $\textit{h-lcc}$ is explained. For each object, consider a cascade of bounding primitives. The more sophisticated primitive is a tighter fit and the simpler one is looser but less expensive. For example, a superquadric (SQ) is a better fit but more expensive than an ellipsoid (E). If the ellipsoid is an upper bound of the superquadric, which means $SQ_i \subseteq E_i$, then when $P(E_1^{g_1}, E_2^{g_1})<\delta$, the $P(SQ_1^{g_1}, SQ_2^{g_1})<\delta$ also holds. In this way, the region where two objects have low collision probability can be quickly screened out if the ellipsoid models do not collide. If not, the more sophisticated model is queried. 

\section{Results}

We first compare the relative goodness of fit using superquadrics and ellipsoids. Then we evaluate the performance, i.e., accuracy and computational time, of the proposed method against existing PCD methods for objects with inaccurate position estimates. In addition, this work assesses the performance of motion planners when using different PCD methods for position errors-only cases. Lastly, a Real2Sim2Real benchmark examines the importance of accounting for pose sensing uncertainty, rather than ignoring the orientation errors of objects in the real-world application. All benchmarks were executed on an Intel Core i9-10920X CPU at 3.5GHz.

\subsection{Evaluation of model fit}
\label{subsec:benchmark:model-fit}
To evaluate the relative goodness of fit of using superquadrics and ellipsoids, we compare their performance for approximating 1) convex polyhedra, 2) cuboids, and 3) cylinders. The fitting algorithm for approximating a polyhedron with a superquadric or an ellipsoid is from an existing work \cite{ruan2022collision}. Given the dimension $[l, w, h]$ of a cuboid, the superquadric approximation is with semi-axes $\mathbf{a}=[l/2, w/2, h/2]$ and epsilons $\bm{\epsilon}=[0.2, 0.2]$. Given the semi-axes $[r_x, r_y]$ and height $h$ of a cylinder, the superquadric approximation is with semi-axes $\mathbf{a}=[r_x, r_y, h/2]$ and epsilons $\bm{\epsilon}=[0.1, 1.0]$. The ellipsoidal approximation for a cuboid and a cylinder is based on the fitting algorithm, which uses the sampled surface points from the object. 
The evaluation metric is 
$$\textit{e}=\frac{V(S_{query}\cap S_{true})}{V(S_{query}\cup S_{true})} \, ,$$ 
where $V(\cdot)$ denotes volume, $S_{query}$ is the model with solid body being evaluated (superquadric or ellipsoid), and $S_{true}$ is the exact object (polyhedron, cuboid, or cylinder). A perfect match would give $e\hspace{-0.2em}=\hspace{-0.2em}1$, and a bad match, such as an overly conservative approximation or one with little overlap, will give a small value of $e$. We generate $N\hspace{-0.2em}=\hspace{-0.2em}100$ random $S_{true}$ and the corresponding candidate model $S_{query}$ and use the average value $\overline{e}=\frac{1}{N} \sum_{i=1}^N e_i$ to assess goodness of fit. 

Results show that: 1) for convex polyhedra, $\overline{e}_{SQ}=0.80189$ and $\overline{e}_{Ellip}=0.66731$; 2) for cuboids, $\overline{e}_{SQ}=0.95703$ and $\overline{e}_{Ellip}=0.67732$; 3) for cylinders, $\overline{e}_{SQ}=0.99396$ and $\overline{e}_{Ellip}=0.71684$. The statistical results show that a superquadric is a tighter representation than an ellipsoid in general cases for convex bodies, especially for commonly used geometric primitives, i.e., cuboids and cylinders.

\subsection{Benchmark on a single query of PCD}
This benchmark compares the proposed PCD methods with state-of-the-art approaches for convex bodies under Gaussian-distributed position errors. Two shape models (ellipsoids and superquadrics) and two error scenarios (one object vs. both objects having independent position errors) yield four test cases: ellipsoids-single-error, superquadrics-single-error, ellipsoids-two-errors, and superquadrics-two-errors.

\subsubsection{Benchmark setting}
Each test generates 100 random object pairs. The semi-axes $[a_1, a_2, a_3]$ of a body are sampled from $(0.2,1.2)$ m. The epsilon variables are randomly sampled in the range of $(0.01, 0.2)$ for superquadrics. Object centers are sampled in $(0.0, 0.1)$ m and $(0.3, 1.3)$ m, respectively. The orientations are uniformly sampled in $\SO(3)$.  Position errors follow a Gaussian distribution with covariance $\Sigma_i=R\Sigma R^T$, where $\Sigma=[4.8,0,0;0,4.8,0;0,0,6.0]*10^{-4}$ is in each object’s local frame and $R\in\SO(3)$. 

\begin{table}[!t]
\caption{Algorithms used in the PCD benchmark.} 
\label{tab:pcd-benchmark-methods}
\resizebox{\linewidth}{!}{%
\begin{tabular}{c c}
\hline  
Notation & Method                                  \\ [0.5ex] \hline 
Baseline for single error & Monte-Carlo Sampling \\
Baseline for two errors \cite{lambert2008fast}         & Fast Monte-Carlo sampling                    \\ 
EB95 \cite{dawson2020provably}        & Bounding volume with $95\%$ of confidence \\
Divergence \cite{park2020efficient}        & Divergence theorem               \\
lcc-center \cite{liu2023tight}  &  Linear chance constraint for ellipsoids \\
\textbf{lcc-tangent (ours)} & (\ref{eq:method:least-squares}) \\
\hline
\end{tabular}}
\end{table}

\resizebox{\columnwidth}{!}{%
\begin{threeparttable}[tb]
\caption{Comparison of PCD methods on accuracy and computation time under position estimation errors.}
\label{tab:PCD-benchmark}
\begin{tabular}{ccccc}
\hline
\multirow{2}{*}{Test setting}
& \multirow{2}{*}{PCD method}
& \multirow{2}{*}{Mean$^3$} 
& \multirow{2}{*}{Variance$^3$}         
& \multirow{2}{*}{\begin{tabular}[c]{@{}c@{}}Computation\\ time(s)\end{tabular}} \\
 & & & & \\ \hline
\multicolumn{1}{c}{}              & EB95                 & 0.0511          & 0.0310          & 0.0069              \\
\multicolumn{1}{c}{ellipsoids}    & Divergence           & 0.0945          & 0.0651          & 0.0044              \\
\multicolumn{1}{c}{single error$^1$}  & lcc-center           & 0.0296          & 0.0096          & \textbf{1.7396e-4}  \\
\multicolumn{1}{c}{}              & \textbf{lcc-tangent (ours)} & \textbf{0.0162} & \textbf{0.0063} & 0.0090              \\ \hline
\multicolumn{1}{c}{}              & EB95                 & 0.4885          & 0.2172          & 0.0093              \\
\multicolumn{1}{c}{superquadrics} & Divergence           & 0.5072          & 0.2114          & 0.0172              \\
\multicolumn{1}{c}{single error$^1$}  & lcc-center           & 0.5634          & \textbf{0.1949}          & \textbf{1.9544e-4}  \\
\multicolumn{1}{c}{}              & \textbf{lcc-tangent (ours)} & \textbf{0.4153} & 0.2099 & 0.0143              \\ \hline
\multicolumn{1}{c}{}              & EB95                 & 0.1435          & 0.1014          & 0.0053              \\
\multicolumn{1}{c}{ellipsoids}    & Divergence           & 0.1041          & 0.0713          & 0.0040              \\
\multicolumn{1}{c}{two errors$^2$}    & lcc-center           & 0.0212          & 0.0055          & \textbf{1.4908e-04} \\
\multicolumn{1}{c}{}              & \textbf{lcc-tangent (ours)} & \textbf{0.0142} & \textbf{0.0043} & 0.0097              \\ \hline
\multicolumn{1}{c}{}              & EB95                 & 0.2439          & 0.1597          & 0.0128              \\
\multicolumn{1}{c}{superquadrics} & Divergence           & 0.1334         & 0.1231        & 0.0193              \\
\multicolumn{1}{c}{two errors$^2$}    & lcc-center           & 0.2078          & 0.1105          & \textbf{2.1014e-04} \\
\multicolumn{1}{c}{}              & \textbf{lcc-tangent (ours)} & \textbf{0.0226} & \textbf{0.0129} & 0.0136              \\ \hline
\end{tabular}%
\begin{tablenotes}
  \small
  \item  ${}^1$ Only object $S_2$ has position estimation error. 
  \item  ${}^2$ Objects $S_1$ and $S_2$ have independent position estimation errors.
  \item  ${}^3$ Mean and variance of the differences (method minus baseline) for each method.
  \item 
\end{tablenotes}
\end{threeparttable}
}

The PCD methods included in the benchmark are summarized in Table~\ref{tab:pcd-benchmark-methods}. 
The baseline for single-error cases employs $10^4$ Monte Carlo samples of translated copies of $S_2$ based on its position error distribution. For double-error cases, a fast Monte Carlo approach generates $10^5$ translated copies of object $S_2$ based on the relative position error distribution \cite{lambert2008fast}. In both cases, the deterministic collision status of translated $S_2$ with $S_1$ is computed for each sample. \textit{EB95} is the enlarged bounding volume method with $95\%$ confidence \cite{dawson2020provably}. 
Divergence \cite{park2020efficient} applies the divergence theorem to meshed object surfaces (100 points for ellipsoids, 1600 for superquadrics), excluding mesh construction time for fairness of comparison.
\textit{lcc-center} uses the linear chance constraint method \cite{zhu2019chance}\cite{liu2023tight}. As for applying \textit{lcc-center} in superquadrics cases, we first compute the minimum volume enclosing ellipsoids for the superquadrics, and the computation time for computing the bounding ellipsoid is not counted for fairness. \textit{lcc-tangent} is the proposed method of this study in (\ref{eq:method:least-squares}). 

\subsubsection{Benchmark results}
The results are summarized in Table~\ref{tab:PCD-benchmark}. Because all PCD methods are approximations of the collision probability, the results can be higher than 1, while a probability by definition cannot exceed $1$. Therefore, if any approximation exceeds 1, we truncate its value to 1. 
To compare how close the approximation of each PCD method is to the baseline result, we compute the mean and variance of the value of each PCD method result minus the ground truth. A good approximation would give mean and variance close to zero.
Overall, \textit{lcc-tangent} is most accurate, closely matching the baseline with lower mean and variance. Compared to \textit{divergence}, which relies on mesh discretization, \textit{lcc-tangent} approximates the collision probability only using one inequality. Although \textit{EB95} can increase accuracy by raising the expansion confidence, this also increases computation time. Meanwhile, \textit{lcc-center} is the fastest since it skips optimization, but it is generally less accurate than \textit{lcc-tangent}, which leverages a plane tangent to both the exact Minkowski sum boundary and the position error's confidence level surface to avoid conservativeness.

\subsection{Benchmark on planning under position estimation errors}
We evaluate how PCD computational accuracy and runtime affect motion planning, comparing \textit{lcc-center} (baseline), \textit{lcc-tangent}, and \textit{h-lcc} used as the sub-modules in RRT-connect (sampling-based) \cite{sucan2012the-open-motion-planning-library} and STOMP (optimization-based) \cite{kalakrishnan2011stomp} planners.

\begin{figure}[tb]
\centering
\includegraphics[width=0.45\textwidth]{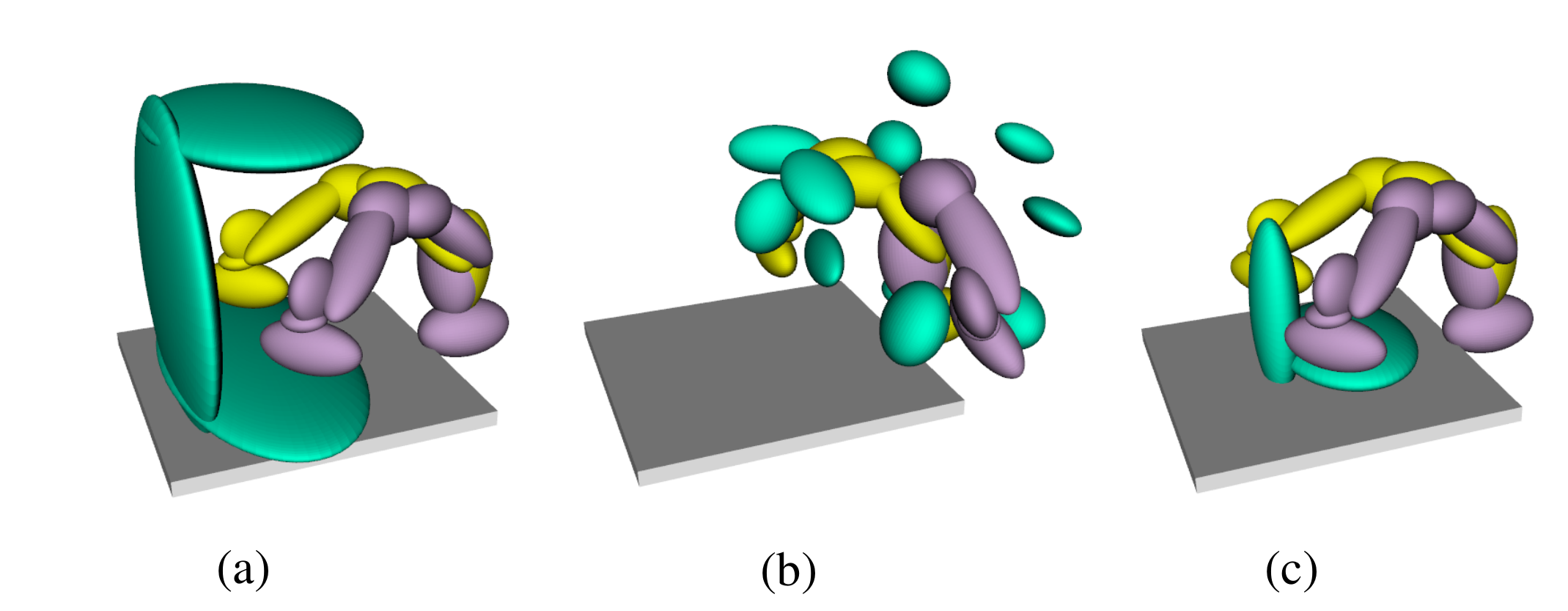}
\caption{Three types of environments used in the planning benchmark: (a) clamp, (b) narrow, and (c) sparse. All obstacles are ellipsoids and are subject to position errors. The violet and yellow colors represent the start and goal joint configurations of the arm. Each robot link is bounded by an ellipsoid.}
\label{fig:results:env-setting}
\end{figure}

\begin{table*}[tb]
\centering
\caption{Performance of motion planners with different PCD methods under position-only uncertainty. }
\label{tab:results:planning-benchmark}
\resizebox{0.8\linewidth}{!}{%
\begin{tabular}{cccccc}
\hline
Motion planner & Environment & PCD method & Planning success rate & Path length (rad)  & Planning time (s) \\ \hline
\multirow{6}{*}{RRTconnect} &  & lcc-center & 100\% & 5.24 &  1.28 \\
 & clamp & \textbf{lcc-tangent (ours)} & 100\% & 4.90  & 2.62  \\
 &  & \textbf{h-lcc (ours)} & 100\% & \textbf{3.72} & \textbf{1.06}  \\ \cline{2-6} 
 &  & lcc-center & 8\% & \textbf{9.08} & \textbf{107.78} \\
 & narrow & \textbf{lcc-tangent (ours)} & \textbf{18\%} & 9.44  & 132.92 \\
 &  & \textbf{h-lcc (ours)}& 10\% & 10.52 & 121.11  \\ \hline
\multirow{3}{*}{STOMP} &  & lcc-center & 100\% & 8.40 & 21.04\\
 & sparse & \textbf{lcc-tangent (ours)} & 100\% & \textbf{7.15} & 17.44 \\
 &  & \textbf{h-lcc (ours)} & 100\% & \textbf{7.15}  & \textbf{13.06} \\ \hline
\end{tabular} }
\end{table*}

\begin{figure*}[!t]
\centering
\includegraphics[width=0.9\linewidth]{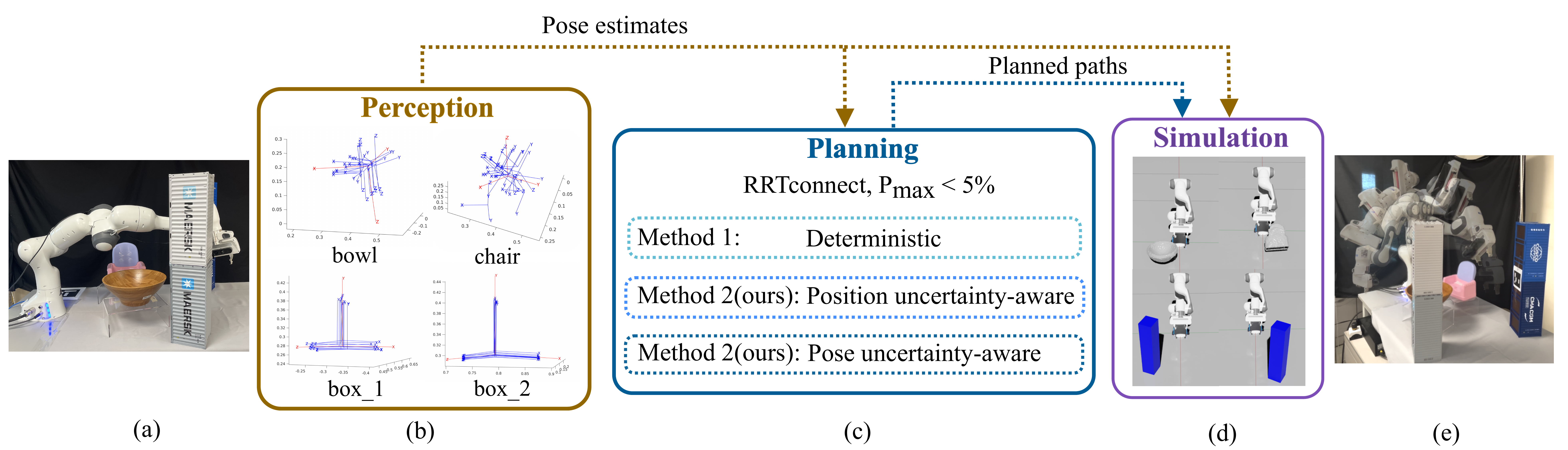}
\caption{Real2Sim2Real pipeline to evaluate the reliability of planned paths using different PCD methods. (a) A typical manipulation setting with four objects. (b) Perception results for each object. (c) Motion planning with different levels of uncertainty awareness. (d) Rolling out the possibility of the robot executing the planned path in simulation (Isaac Sim) and calculating the collision probability in execution. (e) A collision-free path in simulation is executed in the real world, which is planned under considering pose sensing uncertainty.  }
\label{fig:Real2Sim2Real:flow}
\end{figure*}

\subsubsection{Benchmark setting} The environment settings are shown in Fig. \ref{fig:results:env-setting}. The high-precision industrial robot Franka Emika with the fixed basement is used, and the pose estimation error of the robot can be ignored. The start and goal joint configurations are pre-defined for each setting. Each link of the robot arm is bounded by an ellipsoid, and all obstacles are ellipsoids. In addition, the obstacles are only considered to have position errors, where $\Sigma_i=R\Sigma R^T$, $\Sigma=[4.8,0,0;0,4.8,0;0,0,6.0]*10^{-4}$ in the body-fixed frame, and $R\in \SO(3)$ is the obstacle's orientation. Because all obstacles are static, their position errors will not be propagated. We set the collision threshold to be $\delta=0.05$. A robot configuration is invalid if any PCD result for a link–obstacle pair exceeds $\delta$. Each environment setup is tested 100 times.

\subsubsection{Benchmark results} The average path length and planning time for the success cases are listed in Table~\ref{tab:results:planning-benchmark}.

The narrow environment result shows that with more obstacles in the environment, the planning success rates using our methods will be higher than the baseline. This is because the number of valid robot configurations, for which the collision probability between any of its links and objects is lower than the given threshold ($\delta<5\%$), is much lower than in a sparse environment. Our methods, especially the \textit{lcc-tangent}, give a tighter approximation of the collision probability than the baseline and thus can help the planner avoid missing valid configurations.
By avoiding overly conservative collision checks, our methods tend to find a shorter path in the clamp and the sparse scene.  

The observed differences in computation time between the two planners stem from their fundamentally different path-planning approaches. RRTconnect relies on random configuration exploration to find a valid path. Since the query time of \textit{lcc-tangent} is longer than that of \textit{lcc-center}, the overall exploration speed of RRTconnect is slower when using \textit{lcc-tangent}. In contrast, STOMP's iterative process allows it to benefit from the precise collision checking provided by \textit{lcc-tangent} and \textit{h-lcc}. The accuracy of these methods enables STOMP to converge faster to an optimized path, leading to reductions in both path length and planning time.

\subsection{Real2Sim2Real evaluation pipeline}    
This benchmark is designed to test the safety of the planned path in execution when considering different types of sensing uncertainty in motion planning. Here, we create a typical manipulation setting, and the objects' pose estimation is done in the real world. The pose estimation results are used in motion planning when considering no sensing errors (baseline), position estimation errors, and both position and orientation estimation errors. After that, we test the collision probability between the robot and objects in the simulation, where the simulation environment rolls out the possibility of the robot executing the planned path in the real world. 

\subsubsection{Benchmark setting}

The experiment setting is shown in Fig. \ref{fig:Real2Sim2Real:flow} (a).
A Franka Emika robot arm moves from a fixed start to a goal configuration. Each robot link is bounded by a superquadric. Object shapes are known, while their poses are sensed via RGB-D cameras: bowl and chair through iterative closest point (ICP), and boxes through ArUco markers. From each pose estimate, we extract the position error distribution and mean orientation.

We run RRTconnect with three collision checkers over 100 trials each: a deterministic checker (\textit{cfc-dist-ls} \cite{ruan2022collision}), a PCD method handling position errors (\textit{lcc-tangent}), and a PCD method for pose errors (encapsulated surfaces + \textit{lcc-tangent}).

Each set of 100 planned paths is executed four times in simulation, once for each of the four objects. At each run, the object's pose is updated randomly from the pose estimations from real-world measurements. Paths are marked “unsafe” if any via point collides with an object, and the collision probability is the fraction of unsafe paths.

\subsubsection{Benchmark results} 
\begin{table*}[tb]
\centering
\caption{Simulated collision risk of paths generated by motion planners with different uncertainty awareness.}
\label{tab:benchmark:real2sim2real}
\resizebox{1.0\linewidth}{!}{%
\begin{tabular}{ccccc}
\hline
 Collision checker & Planning success rate & Path length (rad)   &  Planning time ($s$)  &  Collision risk in simulation \\ \hline
 Deterministic        & 100\% & 4.48 & 0.49 & 29\%   \\
 \textbf{Position uncertainty-aware$^1$} & 100\% & 4.69 & 2.77271 & 9\% \\ 
 \textbf{Pose uncertainty-aware$^1$} & 100\%  & 4.84 & 9.62  & 2\%   \\ \hline
\multicolumn{2}{c}{$^1$ The maximum collision probability constraint in planning is set to $\delta=5\%$. }
\end{tabular} 
}
\end{table*}
The pose estimations for each object are shown as Fig. \ref{fig:Real2Sim2Real:flow} (b), where the red coordinate is the mean pose, and the blue ones are the pose estimations. ICP yields larger errors for the bowl and chair than ArUco for the boxes. The results are summarized in Table~\ref{tab:benchmark:real2sim2real}. For the position uncertainty-aware PCD method and the pose uncertainty-aware PCD method, if the maximum collision probability approximation between any pair of robotic arm links and objects is larger than the threshold $\delta=5\%$, then the configuration is invalid. The results show that planners using the three types of PCD methods successfully find a path in each call. In principle, the robot should execute the planned path in the real world with the collision risk being zero or below $\delta$. However, the collision probabilities of path execution in the simulation are $29\%$ with no sensing errors, $9\%$ with only position estimation errors, and $2\%$ with both the position and orientation estimation errors considered in motion planning. The high collision risk is obvious when no sensing errors are considered. Interestingly, when not considering the orientation errors, the collision probability in execution is larger than the set threshold of $5\%$ in motion planning. Most non-safe cases happen because the robot arm collides with an object when its orientation in simulation deviates greatly from the mean orientation. Thus, the orientation errors are not ignorable even when the perception method is accurate. By creating an encapsulating surface for the object using (\ref{eq:method:SO3}), most surface points of the rotated copies of objects are encapsulated, which is similar to stretching the original body in deviated orientations. 
The increased planning time when considering the sensing uncertainty is because 1) the single query time of PCD methods is longer than the deterministic method, and 2) more robot configurations in the tree exploration phase are considered invalid.

\section{Conclusion}
This paper introduces an enhanced PCD method with improved accuracy and robustness for robotic motion planning under sensing uncertainty. By leveraging superquadrics for flexible shape representation and incorporating both position and orientation estimation uncertainties, the proposed approach addresses key limitations in existing PCD methods. A hierarchical PCD method (\ie{ \textit{h-lcc}}) further ensures computational efficiency by combining fast screening with precise collision probability approximation where needed. Benchmarks demonstrate that the proposed method is twice as close to the Monte-Carlo sampled baseline as the best existing PCD method and reduces path length by $30\%$ and planning time by $37\%$, respectively.
In addition, the Real2Sim2Real pipeline shows that by considering the orientation errors, the collision probability in executing the planned path is only $2\%$, which is much lower than $9\%$ when considering only the position errors and $29\%$ when ignoring all sensing errors. 
These advancements enable safer and more efficient motion planning for high-DOF robots in cluttered, unstructured environments. 

However, our current framework assumes that the superquadric/ellipsoid models of environmental objects are known a priori and that obstacles are static. In future work, we will integrate online shape approximation to handle previously unseen objects and extend the PCD framework by incorporating trajectory prediction and sensing error propagation to support dynamic obstacles.

\bibliographystyle{IEEEtran}  
\bibliography{main}    
\end{document}